\documentclass[10pt, conference]{IEEEtran}
\IEEEoverridecommandlockouts
\usepackage{amsmath,amssymb,amsfonts}
\usepackage{algorithm}
\usepackage{algpseudocode}
\usepackage{graphicx}
\usepackage{textcomp}
\usepackage[dvipsnames]{xcolor}
\usepackage{tikz}
\usepackage{scalerel}
\usepackage{physics}

\usetikzlibrary{shapes.geometric, arrows, fit, positioning, calc, math, matrix}

\tikzstyle{startstop} = [rectangle, minimum width=0.3cm, minimum height=1cm, text centered]
\tikzstyle{rnode} = [rectangle, minimum width=0.3cm, minimum height=0.5cm, text centered]
\tikzstyle{process} = [rectangle, minimum width=1.3cm, minimum height=1cm, text centered, fill=orange!30]
\tikzstyle{decision} = [diamond, aspect=2, minimum width=1.5cm, minimum height=1cm, text centered, fill=BlueGreen!50]
\tikzstyle{arrow} = [thick,->,>=stealth]
\tikzstyle{tblock} = [rectangle, rounded corners, line width=1.2pt, minimum width=5cm,   minimum height=0.7cm,text centered, fill={rgb,255:red,201; green,202; blue,202}]
\tikzstyle{pblock} = [rectangle, rounded corners, line width=1.3pt, minimum width=1.5cm, minimum height=1.2cm,text centered, fill={rgb,255:red,201; green,222; blue,227}]
\tikzstyle{rblock} = [rectangle, rounded corners, line width=1.3pt, minimum width=1.5cm, minimum height=1.2cm,text centered, fill={rgb,255:red,171; green,200; blue,207}, opacity=0.8]
\tikzstyle{cblock} = [rectangle, rounded corners, line width=1.3pt, minimum width=1.5cm, minimum height=1.2cm,text centered, fill={rgb,255:red,224; green,224; blue,191}]
\tikzstyle{bblock} = [rectangle, rounded corners, line width=1.3pt, minimum width=1.5cm, minimum height=1.2cm,text centered, fill={rgb,255:red,224; green,224; blue,191}, opacity=0.7]
\tikzstyle{wblock} = [rectangle, rounded corners, line width=1.3pt, minimum width=1.5cm, minimum height=0.2cm,text centered]

\def\BibTeX{{\rm B\kern-.05em{\sc i\kern-.025em b}\kern-.08em
    T\kern-.1667em\lower.7ex\hbox{E}\kern-.125emX}}

\begin{document}
\bstctlcite{IEEEexample:BSTcontrol}

\title{Trust-Region Method with Deep Reinforcement Learning in Analog Design Space Exploration}
\author{\IEEEauthorblockN{Kai-En Yang\IEEEauthorrefmark{1}, Chia-Yu Tsai\IEEEauthorrefmark{2}, 
		Hung-Hao Shen\IEEEauthorrefmark{2}, Chen-Feng Chiang\IEEEauthorrefmark{2}, Feng-Ming Tsai\IEEEauthorrefmark{2}, \\
		Chung-An Wang\IEEEauthorrefmark{2}, Yiju Ting\IEEEauthorrefmark{2}, Chia-Shun Yeh\IEEEauthorrefmark{2},
		Chin-Tang Lai\IEEEauthorrefmark{2}}
\IEEEauthorblockA{\IEEEauthorrefmark{1}EECS, National Tsing Hua University, Hsinchu, Taiwan
\IEEEauthorrefmark{2}MediaTek Inc., Hsinchu, Taiwan}}

\IEEEoverridecommandlockouts
\IEEEpubid{\makebox[\columnwidth]{978-1-6654-3274-0/21/\$31.00~\copyright2021 IEEE \hfill} \hspace{\columnsep}\makebox[\columnwidth]{ }}

\maketitle

\IEEEpubidadjcol

\begin{abstract}
	This paper introduces new perspectives on analog design space search.
	To minimize the time-to-market, this endeavor better cast as constraint satisfaction problem than global optimization defined in prior arts.
	We incorporate model-based agents, contrasted with model-free learning, to implement a trust-region strategy. As such, simple feed-forward networks can be trained with supervised learning, where the convergence is relatively trivial.
	Experiment results demonstrate orders of magnitude improvement on search iterations.
	Additionally, the unprecedented consideration of PVT conditions are accommodated.
	On circuits with TSMC 5/6nm process, our method achieve performance surpassing human designers.
	Furthermore, this framework is in production in industrial settings.

\end{abstract}


\section{Introduction}
	The annual increment of computing power described by Moore's law is pioneering unprecedented possibilities.
	This remarkable progress has been accompanied by a collinearity with tremendous increases in chip design complexity. One example of this is the growth in PVT (process, voltage, temperature) corners.

	Despite the majority of the SoC area is occupied by digital circuitry, analog circuits are still essential for the chips to be able to communicate with and sense the rest of the world.
	However, the design effort of the analog counterpart is more onerous due to the heavy requirements of human expertise, and the absence of automation tools similar to digital circuit design.

	One of the labor intensive task in analog design is transistor sizing. Currently, it is mostly done by labouring trial-and-error.
	The designers begin by applying their knowledge about the characteristics of analog circuits and transistors to select a reasonable range of candidate solutions, next explore the space with grid search. Then, get feedback from SPICE circuit simulations.
	Afterward, repeat the procedure until the specifications are met.
	The main challenge of automating sizing is that it has a very large design space, which prior arts suffered from convergence problems and poor scalability \cite{hakhamaneshi2019analog}.

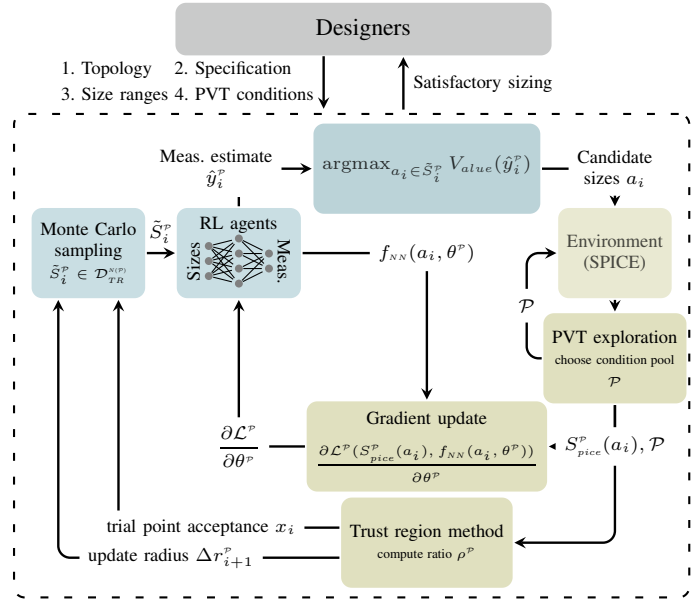
\begin{figure}
\centering
\begin{tikzpicture}[node distance=1cm]
	\tikzstyle{every node}=[font=\small]
	\node(RD)[tblock]{Designers};
	\draw[arrow, line width=0.9pt](RD.south) ++ (-0.53cm, 0.01cm) -- node[anchor=east, align=left]{\scriptsize 1. Topology $\;\,$  2. Specification\\ \scriptsize 3. Size ranges 4. PVT conditions} ++ (0, -0.7cm);
	\draw[arrow, line width=0.9pt](RD.south) ++ (0.53cm, -0.7cm) -- node[anchor=west]{\scriptsize Satisfactory sizing$\;\;\;\;\;\;\;\;\;\;\;\;\;$} ++ (0, 0.7cm);

\end{tikzpicture}

\def\layersep{0.4cm}
\begin{tikzpicture}[node distance=1cm]
	\tikzmath{
		\linewidth = 0.9pt;
	}
	\tikzstyle{every node}=[font=\scriptsize]
	\node(monte)[pblock, align=center]{
		\scriptsize Monte Carlo\\ sampling \\
		\tiny $\tilde{S}_{i}^{\scaleto{\mathcal{P}}{2pt}} \in \mathcal{D}^{\scaleto{N(\mathcal{P})}{2pt}}_{\scaleto{TR}{2pt}}$};
	\node(agent)[pblock, align=center, right of=monte, xshift=1cm, inner sep=2pt]{\scriptsize RL agents\\
		\begin{tikzpicture}[shorten >=1pt, draw=black, node distance=\layersep]
			\tikzmath{
				\neuronsep = 0.2;
			}
			\tikzstyle{neuron}=[circle,fill=black!50,minimum size=3pt,inner sep=0pt]
			\tikzstyle{annot} = [text width=4em, text centered]
			\foreach \name / \y in {1, ..., 3}
				\node[neuron, ] (I-\name) at (0,-\y*\neuronsep) {};
			\foreach \name / \y in {1, ..., 4}
				\path[yshift=\neuronsep*0.5cm] node[neuron] (H-\name) at (\layersep, -\y*\neuronsep cm) {};
			\foreach \name / \y in {1, ..., 2}
				\path[yshift=-\neuronsep*0.5cm] node[neuron, right of=H-3] (O-\name) at (\layersep, -\y*\neuronsep cm) {};
			\foreach \source in {1, ..., 3}
					\foreach \dest in {1, ..., 4}
						\path (I-\source) edge (H-\dest);
			\foreach \source in {1, ..., 4}
				\foreach \dest in {1, ..., 2}
					\path (H-\source) edge (O-\dest);

			\node [rotate=90, rectangle, minimum width=0.3cm, minimum height=0.1cm, text centered, above of=I-2, yshift=-0.2cm] {\scriptsize Sizes};
			\node [rotate=-90, rectangle, minimum width=0.3cm, minimum height=0.1cm, text centered, above of=O-1, yshift=-0.2cm, xshift=0.1cm] {\scriptsize Meas.};

		\end{tikzpicture}
	};

	\node(env)[bblock, right of=agent, xshift = 4cm, align=center]{\scriptsize Environment\\(SPICE)};
	\node(take)[rblock, above=0.5cm of agent.north] at ($(agent)!0.5!(env)$) {$\mathop{\mathrm{argmax}}_{a_{i} \in \tilde{S}^{\scaleto{\mathcal{P}}{2pt}}_{i}} V_{\scaleto{alue}{3pt}}(\hat{y}^{\scaleto{\mathcal{P}}{2pt}}_{i})$};
	\node(pvt)[cblock, below of=env, yshift=-0.4cm, align=center] {PVT exploration\\\tiny choose condition pool\\ \tiny$\mathcal{P}$ };
	\node(grad)[cblock, below=1.95cm of env.south, align=center] at ($(agent)!0.5!(env)$) {\scriptsize Gradient update\\ \\ \tiny $\dfrac{\partial \mathcal{L}^{\scaleto{\mathcal{P}}{2pt}} (S^{\scaleto{\mathcal{P}}{2pt}}_{\scaleto{pice}{3pt}}(a_{i}), f_{\scaleto{NN}{2pt}}(a_{i}, \theta^{\scaleto{\mathcal{P}}{2pt}}))}{\partial \theta^{\scaleto{\mathcal{P}}{2pt}}}$};
	\node(trm)[cblock, below=3.23cm of env.south, align=center] at ($(agent)!0.5!(env)$) {Trust region method\\\tiny compute ratio $\rho^{\scaleto{\mathcal{P}}{2pt}}$};
	\node(pro)[wblock, left=0.4cm of pvt.west, align=center] at ($(pvt)!0.5!(env)$) {$\mathcal{P}$};

	\draw[arrow, line width=\linewidth, rounded corners=1ex](agent.north) |- node[anchor=center, fill=white, align=center, xshift=-0.3cm]{Meas. estimate\\ $\hat{y}^{\scaleto{\mathcal{P}}{2pt}}_{i}$} (take.west);
	\draw[arrow, line width=\linewidth, rounded corners=1ex](take.east) -| node[anchor=center, fill=white, align=center]{Candidate \\sizes $a_{i}$} (env.north);

	\draw[arrow, line width=\linewidth](monte.east) -- node[anchor=center, above]{$\tilde{S}^{\scaleto{\mathcal{P}}{2pt}}_{i}$} (agent.west);

	\draw[arrow, line width=\linewidth, rounded corners=1ex](agent.east) -| node[anchor=center, fill=white]{$f_{\scaleto{NN}{2pt}}(a_{i}, \theta^{\scaleto{\mathcal{P}}{2pt}})$} (grad.north);
	\draw[rounded corners=1ex, line width=\linewidth] (pvt.west) -| (pro.south);
	\draw[rounded corners=1ex, arrow, line width=\linewidth] (pro.north) |- (env.west);
	\draw[arrow, line width=\linewidth] (env) -- (pvt);
	\draw[arrow, line width=\linewidth, rounded corners=1ex](pvt.south) |- (trm.east); 
	\draw[arrow, line width=\linewidth, rounded corners=1ex](pvt.south) |- node[anchor=center, fill=white]{$S^{\scaleto{\mathcal{P}}{2pt}}_{\scaleto{pice}{3pt}}(a_{i}), \mathcal{P}$} (grad.east); 
	\draw[arrow, line width=\linewidth, rounded corners=1ex](grad.west) -| node[anchor=center, fill=white]{$\dfrac{\partial \mathcal{L}^{\scaleto{\mathcal{P}}{2pt}}}{\partial \theta^{\scaleto{\mathcal{P}}{2pt}}}$}(agent.south);

	\draw[arrow, line width=\linewidth, rounded corners=1ex]([yshift=0.2cm]trm.west) -| node[anchor=center, fill=white, xshift=1.1cm]{trial point acceptance $x_{i}$} ([xshift=0.4cm]monte.south); 
	\draw[arrow, line width=\linewidth, rounded corners=1ex]([yshift=-0.2cm]trm.west) -| node[anchor=center, fill=white, xshift=1.5cm]{update radius $\Delta r^{\scaleto{\mathcal{P}}{2pt}}_{i + 1}$} ([xshift=-0.4cm]monte.south);
	\node[draw, thick, loosely dashed, rounded corners, inner xsep=0.6em, inner ysep=0.3em, fit=(monte) (env) (trm) (take)] (box) {};

\end{tikzpicture}
\caption{Framework architecture}
\label{frameworkla}
\end{figure}

	In this paper, we propose a general learning-based search framework (Fig.~\ref{frameworkla}) to help increase R\&D productivity during analog front-end sizing task (Fig.~\ref{flowla}).
	Experiment results demonstrate that our agents can efficiently and effectively accomplish search in state-of-the-art designs with superior performance while achieving area enhancement over human.
	The contributions of our work are severalfold on different levels.
	\begin{itemize}
		\item \textbf{System level} We propose a general framework for IC design space search. The standardized API allows fast migration provided well-formulated problems. The ultimate goal of the framework is to increase R\&D productivity with minimal extra efforts required.
		\item \textbf{Algorithm level} The proposed model-based reinforcement learning (RL) approach directly mimic the dynamics of the SPICE simulation instead of estimating cumulative future reward in model-free learning. This enables a more adequate implementation for usages in the industry.
		\item \textbf{Verification level} Considerations on PVT conditions better suits practical needs. The negligence of PVT conditions on previous works make them far from industrial adaptation.
	\end{itemize}

\section{Prior arts}
	Sizing automation could be framed as an optimization problem.
	Bayesian optimization (BO) is a popular choice because of the sample efficiency in finding the global optimum \cite{lyu2018batch}\cite{zhang2019efficient}.
	Despite of the promising results, the scalability is a major drawback.
	Note that the scalability addressed is the cubical increment of the number of samples, rather than the dimensionality of the space.

	Recently introduced methods primarily leverage the current success in deep learning.
	Model-based RL trained with well-developed supervised learning methods,
	were established as not suitable for IC design space search due to the difficulty in providing sufficient data samples for training \cite{hakhamaneshi2019analog}\cite{tang2018parametric}.
	Some models demand re-training when a new set of specifications is assigned, hence, cannot be reused \cite{Rosa2020}.

	In later publications, model-free RL comes prominence.
	AutoCkt \cite{settaluri2020autockt} aims to train an efficient agent to explore and gain knowledge about the design space. The agents will then be used to generate trajectories during inference.
	Yet, it is rarely necessary to traverse the space in the industry.

	L2DC \cite{wang2018learning} exploits sophisticated sequence-to-sequence modeling using an encoder-decoder technique.
	GCN-RL \cite{wang2020tts} employs the latest innovation - graph convolutional neural networks - to learn features from the structures. This enables the model to reach better transferability between nodes and topologies.
	In spite of their ability to exceed human-level performance, the amount of human-engineering efforts in observation selection, network architecture design, and reward engineering reduce the feasibility of becoming a generalizable automation tool.

\section{Problem formulation}
	Transistor sizing (Fig.~\ref{flowla}) is an iterative process to determine a suitable set of lengths, widths, and multiplicities for each transistor in the topology in order to achieve the desired specifications.
	This scheme is often regarded as a trade-off between constraints. Larger transistor sizes normally lead to greater performance, but consume more power and area.

	Analog circuit sizing can be formulated as a constrained multi-objective optimization problem, defined in \eqref{optim}.
	\begin{equation}
	\begin{aligned}
		\text{Minimize}      \quad & F_{m, c}(X)
		\qquad
		\begin{array}{c}
			m = 1, 2, ... N_{m}\\
			c = 1, 2, ... N_{c}\\
		\end{array}
		\\
		\textrm{subject to}  \quad & C_{d, c}(X)<0
		\qquad
		\begin{array}{c}
			d = 1, 2, ... N_{d}\\
			c = 1, 2, ... N_{c}\\
		\end{array}
		\\
		\quad & X \in \mathcal{D}_{s}
		\label{optim}
	\end{aligned}
	\end{equation}

	where $X$ is a vector of variables to be optimized; $\mathcal{D}_{s}$ is the design space; $F_{c, m}(X)$ is the $m^{th}$ objective function under the $c^{th}$ PVT condition;
	$C(X)_{d, c}$ is the $d^{th}$ constraint under the $c^{th}$ PVT condition.

\section{Proposed framework}
	The proposed framework is shown in Fig.~\ref{frameworkla}.
	Deep RL is believed to be a robust methodology for solving combinatorial search problem in various disciplines without human in the loop, such as games \cite{mnih2015human}\cite{silver2016mastering}\cite{kaiser2019model}, robotic control \cite{lillicrap2015continuous}\cite{levine2016end}, neural architecture search \cite{zoph2016neural}\cite{baker2016designing}, and IC design \cite{sadasivam2018efficient}\cite{zheng2019energy}.
	We thus cast the transistor sizing task as a DRL framework. This allows us to automate the process, while being able to adapt to the environment fast based on past experiences, and evolve over time.
	This system consists of several subsystems described in the following sections.

\subsection{Problem reformulation}
\begin{figure}
\centering
\begin{tikzpicture}[node distance=1cm]
	\tikzstyle{every node}=[font=\scriptsize]
	\node(spec)[startstop]{Spec};
	\node(topo)[process, right of=spec, xshift=0.3cm, align=center]{Topology\\Selection};
	\node(size)[process, right of=topo, xshift=0.8cm, align=center]{Circuit\\ Sizing};
	\node(ver)[decision, right of=size, xshift=1.3cm, align=center]{Verify\\ (SPICE)};
	\node(lay)[process, right of=ver, xshift=1.2cm, align=center]{Layout};
	\draw[arrow](spec) -- (topo);
	\draw[arrow](topo) -- (size);
	\draw[arrow](size) -- (ver);
	\draw[arrow](ver) -- (lay);
	\draw[arrow](ver.north) --node {} ++ (0, 0.2cm) -| (size.north) node[pos=0.25]{} node[pos=0.75] {};
\end{tikzpicture}
\caption{Analog circuit pre-layout design flow}
\label{flowla}
\end{figure}
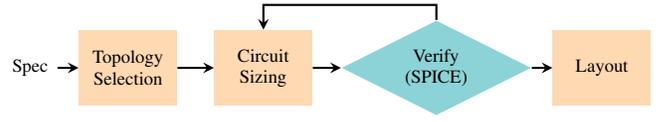

	Transistor sizing (Fig.~\ref{flowla}) has long been formulated as an optimization problem, in which agents and optimizers are implemented to search for optimal points in objective functions.
	However, it is worth a rethink of what is an adequate implementation to integrate with the flow of a designer.
	With the exponential growth in PVT corners during fast technological advances, finding the global optimum is often infeasible. In contrast, meeting the constraints assigned by designers is more practical.

	In this manner, the problem can be reduced to a constraint satisfaction problem (CSP), where the objective function in ~\eqref{optim} is replaced as a 0 constant function.
	More generally, CSP is defined as a triple $\langle X, \mathcal{D}, \mathcal{C} \rangle$ ~\eqref{csp}.
	\begin{equation}
	\begin{aligned}
		X           & = & \{ & x_{1}, x_{2}, ..., x_{n}\}  & \\
		\mathcal{D} & = & \{ & D_{1}, D_{2}, ..., D_{n}\}, & D_{i}           & = & \{ & b_{1}, b_{2}, ..., b_{l}\} \\
		\mathcal{C} & = & \{ & C_{1}, C_{2}, ..., C_{k}\}, & C_{j}           & = & ( & t_{j}, r_{j})
		\label{csp}
	\end{aligned}
	\end{equation}

	where $X$ is a finite set of sizing variables to be searched, each has a non-empty domain $D_{i}$, namely the design space, and $b_{k}$ is the possible value.
	$\mathcal{C}$ is a set of constraints. A constraint is a pair consists of a constraint scope $t_{j}$ and a relation $r_{j}$ over the variables in the scope, limiting feasible permutations of assignments.
	In our case, the relation cannot be explicitly expressed since it is the complex computation inside of the SPICE simulation, denoted as $S_{pice}$ function.

	With this, the intention is to find any complete assignment that is consistent.
	This also prevents over designing the circuit.
	Therefore, an algorithm targeting on fast exploring feasible solutions instead of a global one could make more merit.
	An akin method is meta-learning \cite{lian2019towards}. The model attempts to adapt to new tasks quickly, rather than focusing on a specific environment.

\subsection{Agents} \label{agent}
	Based on the observation and belief that there are multiple satisfactory solutions in different local optima, one effective algorithm for solving such CSP is local search \cite{russell2002artificial}.
	This introduces appealing three-fold advantages:
	1) \textit{Faster environment adaptation}: by reducing the domain to a local region allows fewer iterations for constructing the space. In addition, the circuit space is locally continuous, i.e., neighboring points around a known optimum show similar optimality.
	2) \textit{Model-based agents with supervised learning}: related works criticized supervised learning as per the metric of global goodness of fit. However, it still works given that the local landscape can be captured. Moreover, no reward is involved in the training of model-based agents, making it insensitive to reward engineering.
	3) \textit{Easier implementation and convergence}: model-free agents behave based on a surrogate network modeling the relationship between past trajectories and the next actions to take. Nonetheless, this custom model tends to complicate the problem which makes it hard to converge. Whereas the training routine of supervised learning is relatively trivial.

	All virtues combined actuates a model-based approach, where a direct modeling of the compact circuit space $\mathcal{D}_{L}$ from transistor sizes $X$ to circuit measurements $S_{pice}(X)$ is mapped, imitating the behavior of a SPICE simulator.
	A simple feed-forward neural network $f_{\scaleto{NN}{3pt}}(X; \theta)$ with 3 layers can be used as a SPICE function approximator \eqref{forward}.

\begin{equation}
	\begin{aligned}
		\hat{y} = f_{\scaleto{NN}{3pt}}(X; \theta) \approx S_{pice}(X), X \in \mathcal{D}_{L}
		\label{forward}
	\end{aligned}
\end{equation}
	where $\hat{y}$ is a vector of predicted measurements (e.g., gain, phase margin, etc) w.r.t. a vector of sizes $X$ estimated with weights $\theta$.

	The loss function $J(\theta)$ is simply obtained by the mean squared error (MSE) \eqref{mse}.

\begin{equation}
	\begin{aligned}
		J(\theta) = \dfrac{1}{m} \sum^{m}_{i = 1} (S_{pice}(X^{(i)}) - f_{\scaleto{NN}{3pt}}(X^{(i)}; \theta))^{2}
		\label{mse}
	\end{aligned}
\end{equation}

	Model-based agent is an obscured choice in recent RL solutions. It aims to learn a predictive model $f_{\scaleto{NN}{3pt}}$ to mimic the dynamics of the environment $S_{pice}$, then plan on the model. From this, the number of iterations can be thus reduced.
	A recent novel publication \cite{kaiser2019model} use such agents in Atari game play.
	They claimed that humans can learn fast thanks to the intuitive understanding of the physical processes, and we can apply it to predict the future.
	Therefore, agents who possess such skill could be more sample efficient.

	Each search starts by a random exploration in the design space. Next, the most optimal point will be selected as the local area $\mathcal{D}_{L}$ to dive in.
	The policy is that by modeling the local landscape, a candidate solution can be chosen in the local domain as the next step based on the expected value (discussed in \ref{value}) computed with predicted measurements $\hat{y}$.
	The properties of the local area is granted by the trust region method (TRM), explained in \ref{trm}.
	Even though an optimization or another policy network can be incorporated to find the candidate with the maximum value, to take the advantage of the fast inference time of a neural network, a more vanilla Monte Carlo sampling-based planning is used. The pseudocode of the realization is detailed in Algorithm~\ref{explorer_new}.

\begin{algorithm}
\caption{Fast local explorer algorithm}
	\begin{algorithmic}[1]
	\small
	\State initialize $trajectory \leftarrow [\,]$
	\State initialize $\tilde{S}_{0} \leftarrow ( X^{0}, ..., X^{N} )$, $N$ samples of $X \in \mathcal{D}$
	\State evaluate value $V_{0}$ of $S_{pice}(\tilde{S}_{0})$ via the $V_{alue}$ function
	\State find the best $a_{0} \leftarrow \mathop{\mathrm{argmax}}_{\tilde{S}_{0} \in \mathcal{D}^{N}} V_{0}$ region to search from
		\State initialize $i \leftarrow 0$, $\mathcal{D}^{1}_{\scaleto{TR}{3pt}}$ \& $\theta$ for $f_{\scaleto{NN}{3pt}}$
	\While{not done}
		\State $trajectory.append((a_{i}, S_{pice}(a_{i})))$ for training
		\State $\theta \leftarrow \theta - \alpha \pdv{J(\theta)}{\theta}$
		\State $i \leftarrow i + 1$
		\State sample $m X$s $\tilde{S}_{i} \in \mathcal{D}^{m (i)}_{\scaleto{TR}{3pt}}$, approximate $S_{pice}$ via $f_{\scaleto{NN}{3pt}}(\tilde{S}_{i}; \theta)$
		\State select next $\hat{a}_{i} \leftarrow \mathop{\mathrm{argmax}}_{X \in \tilde{S}_{i}} V_{alue}\circ f_{\scaleto{NN}{3pt}}(\tilde{S}_{i}; \theta)$
		\State $a_{i}, \mathcal{D}^{i + 1}_{\scaleto{TR}{3pt}} \leftarrow TRM(\hat{a}_{i}, S_{pice}(\hat{a}_{i}), \mathcal{D}^{i}_{\scaleto{TR}{3pt}})$
		\If{$a_{i} \in \mathcal{C}$}
			\State $return \: (a_{i}, S_{pice}(a_{i}))$
		\ElsIf{$i > C_{riterion}$}
			\State escape, jump to line $2$
		\EndIf
	\EndWhile

\end{algorithmic}
\label{explorer_new}
\end{algorithm}

\subsection{Trust region method} \label{trm}
	The transition of search space size from a global landscape to a local area is the key factor to the performance of our agents. Thus, the definition of the local properties plays a role in the algorithm efficiency.
	If the compact domain is statically fixed throughout the search, the neural network model could have bad extrapolation properties at the beginning of the episode, when only few samples exist.

	Trust region method defines an iteration-dependent trust region radius $\Delta r_{i}$ where we trust the model $V_{alue} \circ f_{\scaleto{NN}{3pt}}$ to be an adequate representation of the objective function $V_{alue} \circ S_{pice}$.
	At each iteration $i$, a trust region algorithm first solves the trust region subproblem \eqref{trs} to obtain $d^{\star (i)}$.
	In our case, this is realized by Monte Carlo sampling as mentioned in \ref{agent}.

\begin{equation}
	\begin{aligned}
		d^{\star (i)} = \mathop{\mathrm{argmax}}_{X^{i} + d^{i} \in \mathcal{D}^{i}_{\scaleto{TR}{3pt}}} V_{alue} \circ f_{\scaleto{NN}{3pt}} (X^{i} + d^{i}), \\
		\mathcal{D}^{i}_{\scaleto{TR}{3pt}} = \{ X \in \mathcal{D} \mid \left\|X - X^{i}\right\| \leq \Delta r_{i} \}
		\label{trs}
	\end{aligned}
\end{equation}

	where $d^{\star (i)}$ is a vector of optimal trial steps, $\left\|\cdot\right\|$ is a norm, $\mathcal{D}^{i}_{\scaleto{TR}{3pt}}$ is the trust region.

	Then, compute the ratio $\rho^{i}$ of the estimated reduction and the actual reduction. A criterion is set to accept the trial step, i.e. if the ratio is not significant, then the trial point will be rejected.
	Finally, the radius is updated based on the $\rho^{i}$. If the neural network closely approximate the objective function $V_{alue} \circ S_{pice}$, then the trust region will be expanded, shrank otherwise.

\subsection{Reward (value) engineering} \label{value}
	For our agents, there is a value function to estimate the merit of a set of circuit measurements after SPICE simulation.
	This is served as an indication of where to go next.
	Unlike model-free actor-critic methods, values do not participate in training. Consequently, the design of the formula does not affect the convergence of the model.

	In the spirit of simplicity and generalization, we utilize a naive tactic where the value is the sum of normalized measurements. This way, no extra information is acquired.
	However, in terms of the trade-off between constraints, a second-stage value function could be implemented to explicitly encode the importance of each measurement once the agent enters a optimal local area.

\subsection{PVT exploration strategy}
	In order to push the system to production, one important aspect to consider is the PVT conditions.
	To guarantee that a chip is able to work under the variations of fabrications, power supplies, and environments, a number of corners have to be signed off before tape-out.
	Nonetheless, to best of our knowledge, no prior art has accommodated such regime.

	A simple way to explore the conditions would be to test all PVTs every time a new set of assignments is obtained.
	Yet for deployment, this strategy would pose a waste in computing resources and EDA (electronic design automation) tool licenses if the agent is not yet ready for verification.

\begin{figure}
\centering
\begin{tikzpicture}[node distance=0.7cm]
	\tikzset{three sided/.style={
		draw = none,
		append after command={
			[shorten <= -0.5\pgflinewidth]
			([shift={(-1.5\pgflinewidth,-0.5\pgflinewidth)}]\tikzlastnode.north east) edge([shift={( 0.5\pgflinewidth,-0.5\pgflinewidth)}]\tikzlastnode.north west)
			([shift={( 0.5\pgflinewidth,-0.5\pgflinewidth)}]\tikzlastnode.north west) edge([shift={( 0.5\pgflinewidth,+0.5\pgflinewidth)}]\tikzlastnode.south west)
			([shift={( 0.5\pgflinewidth,+0.5\pgflinewidth)}]\tikzlastnode.south west) edge([shift={(-1.0\pgflinewidth,+0.5\pgflinewidth)}]\tikzlastnode.south east)
			}
		}
	}
	\tikzset{two sided/.style={
		draw = none,
		append after command={
			[shorten <= -0.5\pgflinewidth]
			([shift={(-1.5\pgflinewidth,-0.5\pgflinewidth)}]\tikzlastnode.north east) edge([shift={( 0.5\pgflinewidth,-0.5\pgflinewidth)}]\tikzlastnode.north west)
			([shift={( 0.5\pgflinewidth,+0.5\pgflinewidth)}]\tikzlastnode.south west) edge([shift={(-1.0\pgflinewidth,+0.5\pgflinewidth)}]\tikzlastnode.south east)
			}
		}
	}
	\tikzset{end side/.style={
		draw = none,
		append after command={
			[shorten <= -0.5\pgflinewidth]
			([shift={(-1.5\pgflinewidth,-0.5\pgflinewidth)}]\tikzlastnode.north east) edge([shift={( 0.5\pgflinewidth,-0.5\pgflinewidth)}]\tikzlastnode.north west)
			([shift={( 0.5\pgflinewidth,-0.5\pgflinewidth)}]\tikzlastnode.north east) edge([shift={( 0.5\pgflinewidth,+0.5\pgflinewidth)}]\tikzlastnode.south east)
			([shift={( 0.5\pgflinewidth,+0.5\pgflinewidth)}]\tikzlastnode.south west) edge([shift={(-1.0\pgflinewidth,+0.5\pgflinewidth)}]\tikzlastnode.south east)
			}
		}
	}
	\tikzstyle{every node}=[font=\tiny]
	\tikzstyle{nonblock} = [draw, rectangle, line width=0.6pt, minimum width=0.6cm, minimum height=0.4cm, text centered]
	\tikzstyle{nopblock} = [draw, rectangle, line width=0.6pt, minimum width=0.3cm, minimum height=0.4cm, text centered]
	\tikzstyle{redblock} = [draw, rectangle, line width=0.6pt, minimum width=0.3cm, minimum height=0.4cm, text centered, fill=red]
	\tikzstyle{greblock} = [draw, rectangle, line width=0.6pt, minimum width=0.3cm, minimum height=0.4cm, text centered, fill={rgb,255:red,0; green,153; blue,0}]
	\tikzstyle{commentblock} = [rectangle, line width=0.6pt, minimum width=0.3cm, minimum height=0.4cm, text centered, align=center, yshift=-0.2cm]
	\tikzstyle{hitblock} = [rectangle, line width=0.6pt, minimum width=0.3cm, minimum height=0.4cm, text centered, align=center, text=red]

	\tikzmath{
		\blocksep = 0.4;
		\layersep = -0.4cm;
	}

	\foreach \name / \y in {1, ..., 9} \node[nonblock] (a-\name)  at (0, -\y*\blocksep) {PVT \name};
	\foreach \row / \x in {2, ..., 19}
		\foreach \col / \y in {1, ..., 9}
			\node[nopblock, two sided] (n-\row-\col) at (\x*0.3+0.35, -\y*\blocksep) {};
	\foreach \col / \y in {1, ..., 9} \node[nopblock, three sided] (n-1-\col) at (1*0.3+0.35, -\y*\blocksep) {};
	\foreach \col / \y in {1, ..., 9} \node[nopblock, end side] (n-20-\col) at (20*0.3+0.35, -\y*\blocksep) {};

	\node[commentblock, above of=n-4-1]  (comment1) at ($(n-1-1)!0.5!(n-7-1)$){\scriptsize Train};
	\node[commentblock, above of=n-8-1, text = blue]  (comment2)  {\scriptsize Test\\ \scriptsize all};
	\node[commentblock, above of=n-12-1]  (comment1) at ($(n-9-1)!0.5!(n-14-1)$) {\scriptsize Train};
	\node[commentblock, above of=n-15-1, text = blue] (comment3) {\scriptsize Test\\ \scriptsize all};
	\node[commentblock, above of=n-17-1]  (comment1) at ($(n-16-1)!0.5!(n-19-1)$)  {\scriptsize Train};
	\node[commentblock, above of=n-20-1, text = blue] (comment4) {\scriptsize Test\\ \scriptsize all};
	\foreach \row / \x in {1, ..., 7}
	{
		\ifnum \row=7 \node[greblock] (n-\row-3) at (\x*0.3+0.35, -3*\blocksep) {};
		\else \node[redblock] (n-\row-3) at (\x*0.3+0.35, -3*\blocksep) {};
		\fi
	}
	\foreach \row / \x in {9, ..., 12}
	{
		\ifnum \row=10 \node[greblock] (n-\row-3) at (\x*0.3+0.35, -3*\blocksep) {};
		\else \node[redblock] (n-\row-3) at (\x*0.3+0.35, -3*\blocksep) {};
		\fi
	}
	\foreach \row / \x in {13, ..., 14} \node[greblock] (n-\row-3) at (\x*0.3+0.35, -3*\blocksep) {};
	\foreach \row / \x in {9, ..., 10} \node[redblock] (n-\row-3) at (\x*0.3+0.35, -6*\blocksep) {};
	\foreach \row / \x in {11, ..., 14}
	{
		\ifnum \row=13 \node[redblock] (n-\row-6) at (\x*0.3+0.35, -6*\blocksep) {};
		\else \node[greblock] (n-\row-6) at (\x*0.3+0.35, -6*\blocksep) {};
		\fi
	}
	\foreach \row / \x in {16, ..., 19}
	{
		\ifnum \row=19 \node[greblock] (n-\row-3) at (\x*0.3+0.35, -3*\blocksep) {};
		\else \node[greblock] (n-\row-3) at (\x*0.3+0.35, -3*\blocksep) {};
		\fi
	}
	\foreach \row / \x in {16, ..., 19}
	{
		\ifnum \row=17 \node[redblock] (n-\row-6) at (\x*0.3+0.35, -6*\blocksep) {};
		\else \node[greblock] (n-\row-6) at (\x*0.3+0.35, -6*\blocksep) {};
		\fi
	}
	\foreach \row / \x in {16, ..., 19}
	{
		\ifnum \row=19 \node[greblock] (n-\row-9) at (\x*0.3+0.35, -9*\blocksep) {};
		\else \node[redblock] (n-\row-9) at (\x*0.3+0.35, -9*\blocksep) {};
		\fi
	}

	\foreach \col / \y in {1, ..., 2} \node[greblock] (n-8-\col) at (8*0.3+0.35, -\y*\blocksep) {};
	\foreach \col / \y in {4, ..., 7}
	{
		\ifnum \col=4
			\node[greblock] (n-8-\col) at (8*0.3+0.35, -\y*\blocksep) {};
		\else
			\node[redblock] (n-8-\col) at (8*0.3+0.35, -\y*\blocksep) {};
		\fi
	}
	\foreach \col / \y in {8, ..., 9}
	{
		\ifnum \col=8
			\node[greblock] (n-8-\col) at (8*0.3+0.35, -\y*\blocksep) {};
		\else
			\node[redblock] (n-8-\col) at (8*0.3+0.35, -\y*\blocksep) {};
		\fi
	}
	\foreach \col / \y in {1, ..., 2} \node[greblock] (n-15-\col) at (15*0.3+0.35, -\y*\blocksep) {};
	\foreach \col / \y in {4, ..., 5} \node[greblock] (n-15-\col) at (15*0.3+0.35, -\y*\blocksep) {};
	\foreach \col / \y in {7, ..., 8} \node[greblock] (n-15-\col) at (15*0.3+0.35, -\y*\blocksep) {};
	\node[redblock] (n-15-9) at (15*0.3+0.35, -9*\blocksep) {};

	\foreach \col / \y in {1, ..., 2} \node[greblock] (n-20-\col) at (20*0.3+0.35, -\y*\blocksep) {};
	\foreach \col / \y in {4, ..., 5} \node[greblock] (n-20-\col) at (20*0.3+0.35, -\y*\blocksep) {};
	\foreach \col / \y in {7, ..., 8} \node[greblock] (n-20-\col) at (20*0.3+0.35, -\y*\blocksep) {};

	\path [draw=blue, line width = 1.5pt] (n-8-1.north west) -- (n-8-9.south west);
	\path [draw=blue, line width = 1.5pt] (n-8-1.north east) -- (n-8-9.south east);
	\path [draw=blue, line width = 1.5pt] (n-15-1.north west) -- (n-15-9.south west);
	\path [draw=blue, line width = 1.5pt] (n-15-1.north east) -- (n-15-9.south east);
	\path [draw=blue, line width = 1.5pt] (n-20-1.north west) -- (n-20-9.south west);
	\path [draw=blue, line width = 1.5pt] (n-20-1.north east) -- (n-20-9.south east);

	\path [arrow, draw=black, line width = 0.6pt] ([yshift=-0.2cm]$(n-14-3.north west)!0.5!(n-14-3.north east)$) -- ([yshift=0.28cm]$(n-14-4.south east)!0.5!(n-14-4.south west)$);
	\path [arrow, draw=black, line width = 0.6pt] ([yshift=0.2cm]$(n-14-6.south west)!0.5!(n-14-6.south east)$) -- ([yshift=-0.1cm]$(n-14-5.north east)!0.5!(n-14-5.north west)$);
	\node [hitblock] (hit1) at (14*0.3+0.1, -5*\blocksep+0.33) {Hit spec};

	\path [arrow, draw=black, line width = 0.6pt] ([yshift=-0.2cm]$(n-7-3.north west)!0.5!(n-7-3.north east)$) -- ([yshift=0.28cm]$(n-7-4.south east)!0.5!(n-7-4.south west)$);
	\node [hitblock] (hit2) at (7*0.3+0.1, -5*\blocksep+0.33) {Hit spec};

	\node [hitblock, left of=n-8-6, xshift=-0.2cm] (hit3) {Worst cond.};
	\path [arrow, draw=black, line width = 0.6pt] ([yshift=-0.2cm]$(n-8-6.north west)!0.5!(n-8-6.north east)$) -- ([xshift=-0.2cm]$(n-8-6.north west)!0.5!(n-8-6.south west)$);

\end{tikzpicture}
\centering
\caption[caption]{Progressive PVT exploration strategy(block: 1 EDA time; red: not meet spec; green: meet spec)}
\label{PVT_tikz}
\end{figure}
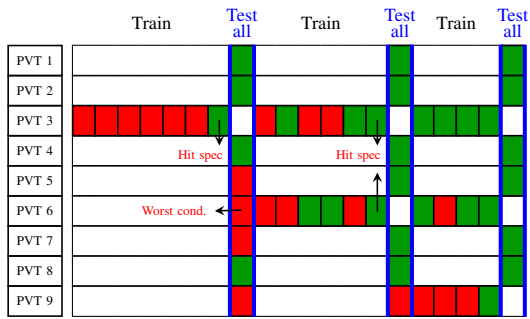

	Inspired by heuristics of designers, first focus the search on a single condition, usually the most difficult condition to search, assuming that by overcoming the hardest PVT, easier ones would be easy.
	Once all the specifications are met, verifications will be conducted to confirm that the set of assignments are legal under all other conditions as well.
	Accordingly, the proposed progressive strategy is that if the initial condition does not apply, the corner with the least value will be chosen to be searched next, until all constraints are satisfied under all conditions (Fig.~\ref{PVT_tikz}).
	If multiple corners are under search, the candidate solution will be taken as the complete assignments with the lowest expected value.
	Note that each PVT condition has its own independent model.

	In Algorithm.~\ref{explorer_new}, the search is demonstrated in an uniform condition. However, it is painless to generalize to our progressive strategy.

\subsection{API}
	A framework with a powerful search algorithm could be wasteful if the user interface is unfriendly or requires extra effort to use.
	In the introduced API, only information that the designers need in their original flow is acquired.
	In particular, human experts will need to identify the transistor sizes to tune, the ranges of variables, the circuit topology, the measurements to observe from SPICE simulations, and the specifications for each corner.
	Once the configuration is set, an automatic script will construct necessary components, namely the neural network architectures and hyperparameters of the network, which are also dynamically scheduled on the fly.
	That is to say, our framework is equivalent to a SPICE decorator where the DRL agents are encapsulated, which seamlessly automates transistor sizing with generalization.

\section{Experiment results}

\subsection{Experimental setup}

	Our experiments are conducted on both academic and industrial settings to evaluate the feasibility and the capability of the DRL agents.

	During the development phase, the agents are tested on a two-stage opamp with BSIM 45/22nm processes simulated on an open sourced NGSPICE simulator developed by UC Berkeley.
	Two scenarios that are often encountered in practice, specifically, process porting and PVT exploration are analyzed.

	To demonstrate that the proposed framework is beneficial in production, we cooperate with industrial professional designers and evaluate the agents on two industrial cases with the most advanced TSMC 5nm and 6nm technologies.
	Simulations are conducted in Cadence Spectre.

\subsection{45nm two-stage opamp}
	First, we benchmark our method with various baseline models including random search,
	customized BO (designed, implemented, and tuned by ourselves),
	Advantage Actor Critic (A2C) \cite{mnih2016asynchronous},
	Proximal Policy Optimization (PPO) \cite{schulman2017proximal}, and
	Trust Region Policy Optimization (TRPO) \cite{schulman2015trust} implemented by Stable-Baselines \cite{stable-baselines}.
	The customized BO utilizes dynamic balancing of exploration \& exploitation, and also substitutes Gaussian Process with extra-tree regressor.
	All model-free agents follow the same observation design in AutoCkt \cite{settaluri2020autockt}, and use the same reward function as our agents.
	This experimental environment is set on a BSIM 45nm process two-stage opamp simulated in a single PVT with a design space size of $10^{14}$. The comparison is shown in Table~.\ref{opamp45}.

	Each experiment has a 10k-step limitation.
	For the customized BO and our model-based agents, 100 experiments are ran to prove the stability, whereas for agents in stable-baselines, only 10 are executed since it would have taken about a month to complete.

\begin{table}[htbp]
\caption{Performance of agents in 45nm two-stage opamp}
\begin{center}
\begin{tabular}{c|ccc}
	\hline
	                                       & Success rate & Average iterations \\ \hline
	Random search                          & 100\%        & 8565               \\ \hline
	Customized BO                          & 100\%        & 330                \\ \hline
	A2C \cite{mnih2016asynchronous}        & 90\%         & 34797              \\ \hline
	PPO \cite{schulman2017proximal}        & 40\%         & 31503              \\ \hline
	TRPO \cite{schulman2015trust}          & 20\%         & 16350              \\ \hline
	Our method                             & 100\%        & 36                 \\ \hline
\end{tabular}
\end{center}
\label{opamp45}
\end{table}

	The experiments show that random search is a strong baseline in which model-free agents (A2C, PPO, TRPO) failed to reach its performance.
	The reason is that there exists a trade-off between gain and phase margin, which are two constraints of opamp.
	Circuits with high gains often come with unstable phase margins.
	Hence, if the reward formula of model-free agents do not encode such information, the agent can be easily stuck in a local maxima where this happens.

	The customized BO states a solid stability among our experiments, however, the nature to estimate the global distribution gains iterations compared with our local model-based agents.
	As for model-based agents, all specifications can be met within 36 steps on average. Moreover, the standard deviation is 16 steps, which indicates that the search is stable.
	This comparison demonstrates that our idea of implementing an agent good at local exploration is feasible, and can outperform model-free agents by orders of magnitude.

\subsection{Process porting}
	Many circuit topologies have to go through a process migration. To avoid reinventing the wheels every time a new process node is applied, AIP (analog intellectual property) reuse is an important topic worth discussing.
	To better understand how the information obtained from the previous nodes can help speeding up the sizing of the new node, a migration from BSIM 45nm to 22nm process is prepared.
	Each experiment is ran for 100 times in the BSIM 22nm circuit to account for the randomness.

\begin{table}[htbp]
	\caption{Results of process porting from 45nm to 22nm}
	\begin{tabular}{c|ccc}
		\hline
		& Average steps & Min steps & Max Steps \\ \hline
		\begin{tabular}[c]{@{}c@{}}Baseline (random weights,\\ random starting points)$^{\mathrm{a}}$\end{tabular} 			   & 50.17 & 15 & 191 \\ \hline
		\begin{tabular}[c]{@{}c@{}}Weight sharing,\\ starting point sharing$^{\mathrm{b}}$\end{tabular}         & 29.22 & 3  & 310 \\ \hline
		\begin{tabular}[c]{@{}c@{}}Random weights,\\ starting point sharing$^{\mathrm{c}}$\end{tabular}          & 20.74 & 2  & 88  \\ \hline
		\multicolumn{4}{l}{$^{\mathrm{a}}$Directly deploy our method on 22nm without process porting} \\
		\multicolumn{4}{l}{$^{\mathrm{b}}$Use the optimal network weights and optimal solution as starting} \\
		\multicolumn{4}{l}{points that the model found in 45nm process} \\
		\multicolumn{4}{l}{$^{\mathrm{c}}$Apply random weight initialization but start the agent on different} \\
		\multicolumn{4}{l}{optimal solutions found in 45nm process} \\

	\end{tabular}
	\label{porting}
\end{table}

	All three strategies can find solutions 100\% of the time, so the success rate is omitted. 
	Table.~\ref{porting} shows that optimal points from previous nodes are reliable. However, the results state impotent network weights transferability, implying that the distributions between processes are distinctive.
	Interestingly, this phenomenon matches designers' experiences: previously sized circuits are strong references, but the equations describing the physics of transistors could be distinguishing.

\subsection{PVT exploration}
	Finding a sufficient set of sizes in a single condition is only a part of the story. Thus, we test PVT exploration strategies on our method using the two-stage opamp with BSIM 22nm process. The results are illustrated in Table.~\ref{multipvt}.

\begin{table}[htbp]
	\caption{Comparison of PVT exploration strategies}
	\begin{center}
	\begin{tabular}{c|ccc}
		\hline
	                   	            & Average steps    & Min steps & Max steps \\ \hline
		Random search               & failed (10,000+) & NA        & NA        \\ \hline
		Brute force (test all cond.)& 359.4            & 36        & 1305      \\ \hline
		Progressive (random cond.)  & 89.52            & 20        & 450       \\ \hline
		Progressive (hardest cond.) & 72.60            & 15        & 279       \\ \hline
	\end{tabular}
	\end{center}
	\label{multipvt}
\end{table}

	Only random search cannot accomplish the task, other strategies can finish 100\% of the time.
	A 4x improvement of progressive search over brute force search (test all conditions in every iteration) is exemplified.
	An intriguing finding is that while starting from the most difficult condition does make a difference, choosing a random corner to start also produce comparable results.
	This suggests that the progressive search is not sensitive to the initial condition, which is positive for cases where the toughest PVT corner is unidentifiable owing to the number of permutations. 

\subsection{Industrial cases}
	\textbf{LDO (Low-Drop regulator) on TSMC 6nm process} The first industrial example is a LDO with TSMC 6nm process. In this case, the design space possesses about $10^{29}$ possible combinations.

	The number of iterations of human designers is untraceable. Nevertheless, our agent achieved the specification in all corners within 2609 iterations, which is considered fast in designers' opinion.
	Furthermore, the performance obtained surpass designers while producing even lower area, shown in table.~\ref{RL_RD}.
	An interesting discovery is that even some of the device sizes decided by both AI and human are the same, AI still managed to illustrate an area enhancement.

	In comparison with our customized BO, it cannot satisfy all the constraints within a reasonable time. However, the best parameters searched are very close to the specifications.

\begin{table}[htbp]
	\centering
	\caption{Benchmark of LDO circuit sizing with different agents}
	\begin{tabular}{c|ccc}
		\hline
		                                                                     & \# iterations & Loop gain   & Area           \\ \hline
		Specification														 & -			 & $>$40.0dB   & $<$650nm$^{2}$ \\ \hline
		\begin{tabular}[c]{@{}c@{}}Human                     \end{tabular}   & untraceable   & 38.0dB      & 650nm$^{2}$    \\ \hline
		\begin{tabular}[c]{@{}c@{}}Customized BO$^{\mathrm{a}}$\end{tabular} & failed        & 38.2dB      & 604nm$^{2}$    \\ \hline
		\begin{tabular}[c]{@{}c@{}}Our method            \end{tabular}       & 2609          & 40.0dB      & 632nm$^{2}$    \\ \hline
		\multicolumn{4}{l}{$^{\mathrm{a}}$Customized BO did not satisfy the constraints (not shown here), } \\
		\multicolumn{4}{l}{$\quad$ however, it gives very close results} \\
	\end{tabular}
	\label{RL_RD}
\end{table}

	\textbf{ICO (Current-Controlled oscillator) on TSMC 5nm process} An ICO is tested as the second case with TSMC 5nm process. The design space size is $20^{4}$.

	The AI-sized ICO realized a performance on a par with human designers.
	A comparable result is also exhibited in the customized BO. Yet, as mentioned before, the global search strategy cause 4.5 times more iterations than our local search algorithm (Table.~\ref{RL_ICO}).

\begin{table}[htbp]
	\caption{Benchmark of ICO circuit sizing with different agents}
	\begin{center}
	\begin{tabular}{c|ccccc}
		\hline
				        & \# iterations   & Phase noise     & Frequency  \\ \hline
		Specification   & -               & $<$-71dB        & $>$8GHz    \\ \hline
		Human           & untraceable     & -73.31dB        & 8.45GHz    \\ \hline
		Customized BO   & 194             & -72.17dB        & 8.87GHz    \\ \hline
		Our method      & 43              & -71.76dB        & 9.18GHz    \\ \hline
	\end{tabular}
	\end{center}
	\label{RL_ICO}
\end{table}

	Although the design space of the second case is relatively small, this leads to a declaration.
	Numerous concerns in the community are raised on AI safety \cite{leike2017ai}. As an initial assessment, the ability and the characteristics of the designed circuits are unfamiliar.
	Therefore, to ensure that the agents act as intended and to secure the safety of the products, designers have to fix a subset of the parameters, only letting the agents to search for the rest.
	To that end, not until we have a comprehensive rule for regulating the agents can we unlock the full capability of AI.
	Thus in our evaluations, designers went through a rigorous screening process to examine the designed circuits. Fortunately, the sized circuits are ready to tape out.

	From both cases described earlier, the outcome is akin to what is addressed in SimPLe \cite{kaiser2019model}. Transistor sizing is one of the task that benefits from the sample-efficient advantage of model-based agents.

	In summary, the results state the contributions of this framework in two ways:
	\begin{itemize}
		\item provides better performance within a reasonable time
		\item automates the process, leaving human intervention while obtaining granting satisfactory performance
	\end{itemize}

	This also indicates that the model is generalizable across different circuit schematics and process nodes.
	The term generalization here does not refer to network weight level transfer as the convention in machine learning genre, but rather algorithm architecture level.

\section*{Discussion}

	As the results of this work shows, the proposed search algorithm can achieve automation at human level in analog block circuit sizing.
	Even though some sections of the flow could be accomplished by AI agents, but AI would not be useful if the upstream tasks such as system-level architecture design and circuit-level topology selection were not carefully executed by human experts with their rich experiences and knowledges.

	In concern with the real capability of the scalability, certainly, AI is not trained, nor designed to achieve full analog system circuit design since it is computationally infeasible.
	Rather we can embrace current limitations by divide-and-conquer.
	One can feed an appropriate amount of design to our search framework, and by recursion, we can avoid such scalability problem.
	Once again, it is the merit of designers to come up with a segmentation that allows us to leverage the current AI technology.

\section*{Conclusions}
	In this work, we propose a generalizable search framework using learning-based algorithms for solving the analog circuit sizing problem.
	We take a novel direction where a trust-region method is adopted using model-based agents trained with supervised learning.
	This enables fast design space adaptation. Moreover, a PVT exploration strategy is also proposed to account for different working conditions, which is not considered in previous works.

	Practical evaluations on industrial products with advanced TSMC 5/6nm process shows exceptional results. Our agent achieves performance beyond human level while producing smaller area.
	Furthermore, the presented framework is not limited to this specific stage of the flow. Any section that could be cast as a search problem can be transferred and leverage the assistance of this DRL agents.

\bibliographystyle{IEEEtran}
\bibliography{references}

\end{document}